\documentclass[sigconf]{acmart}
\AtBeginDocument{%
  }

\copyrightyear{2025}
\acmYear{2025}
\setcopyright{acmlicensed}\acmConference[MM '25]{Proceedings of the 33rd ACM International Conference on Multimedia}{October 27--31, 2025}{Dublin, Ireland}
\acmBooktitle{Proceedings of the 33rd ACM International Conference on Multimedia (MM '25), October 27--31, 2025, Dublin, Ireland}
\acmDOI{10.1145/3746027.3754952}
\acmISBN{979-8-4007-2035-2/2025/10}

\usepackage{booktabs}
\usepackage{multirow}
\usepackage{makecell}
\usepackage{graphicx}
\usepackage{subcaption}

\begin{document}

\title{Clustering-Oriented Generative Attribute Graph Imputation}

\author{Mulin Chen}
\affiliation{%
	\institution{School of Artificial Intelligence, OPtics and ElectroNics (iOPEN)}
	\city{Northwestern Polytechnical University, Xi'an}
	\country{China}}
\email{chenmulin@nwpu.edu.cn}

\author{Bocheng Wang}
\affiliation{%
	\institution{School of Artificial Intelligence, OPtics and ElectroNics (iOPEN)}
	\city{Northwestern Polytechnical University, Xi'an}
	\country{China}}
\email{wangbocheng@mail.nwpu.edu.cn}

\author{Jiaxin Zhong}
\affiliation{%
	\institution{School of Artificial Intelligence, OPtics and ElectroNics (iOPEN)}
	\city{Northwestern Polytechnical University, Xi'an}
	\country{China}}
\email{jiaxinzhong66@gmail.com}

\author{Zongcheng Miao}
\affiliation{%
	\institution{School of Artificial Intelligence, OPtics and ElectroNics (iOPEN)}
	\city{Northwestern Polytechnical University, Xi'an}
	\country{China}}
\email{miaozongcheng@nwpu.edu.cn}

\author{Xuelong Li}
\authornote{Corresponding author.}
\affiliation{%
	\institution{Institute of Artificial Intelligence (TeleAI)}
	\city{China Telecom, Beijing}
	\country{China}}
\email{xuelong\_li@ieee.org}

\renewcommand{\shortauthors}{Mulin Chen, Bocheng Wang, Jiaxin Zhong, Zongcheng Miao, and Xuelong Li}

\begin{abstract}
  Attribute-missing graph clustering has emerged as a significant unsupervised task, where only attribute vectors of partial nodes are available and the graph structure is intact. The related models generally follow the two-step paradigm of imputation and refinement. However, most imputation approaches fail to capture class-relevant semantic information, leading to sub-optimal imputation for clustering. Moreover, existing refinement strategies optimize the learned embedding through graph reconstruction, while neglecting the fact that some attributes are uncorrelated with the graph. To remedy the problems, we establish the Clustering-oriented Generative Imputation with reliable Refinement (CGIR) model. Concretely, the subcluster distributions are estimated to reveal the class-specific characteristics precisely, and constrain the sampling space of the generative adversarial module, such that the imputation nodes are impelled to align with the correct clusters. Afterwards, multiple subclusters are merged to guide the proposed edge attention network, which identifies the edge-wise attributes for each class, so as to avoid the redundant attributes in graph reconstruction from disturbing the refinement of overall embedding. To sum up, CGIR splits attribute-missing graph clustering into the search and mergence of subclusters, which guides to implement node imputation and refinement within a unified framework. Extensive experiments prove the advantages of CGIR over state-of-the-art competitors.
\end{abstract}




\maketitle

\section{Introduction}

\label{sec:intro}

\begin{figure}[t]
	\centering
	\begin{subfigure}{0.95\linewidth}
		\includegraphics[width=1\linewidth]{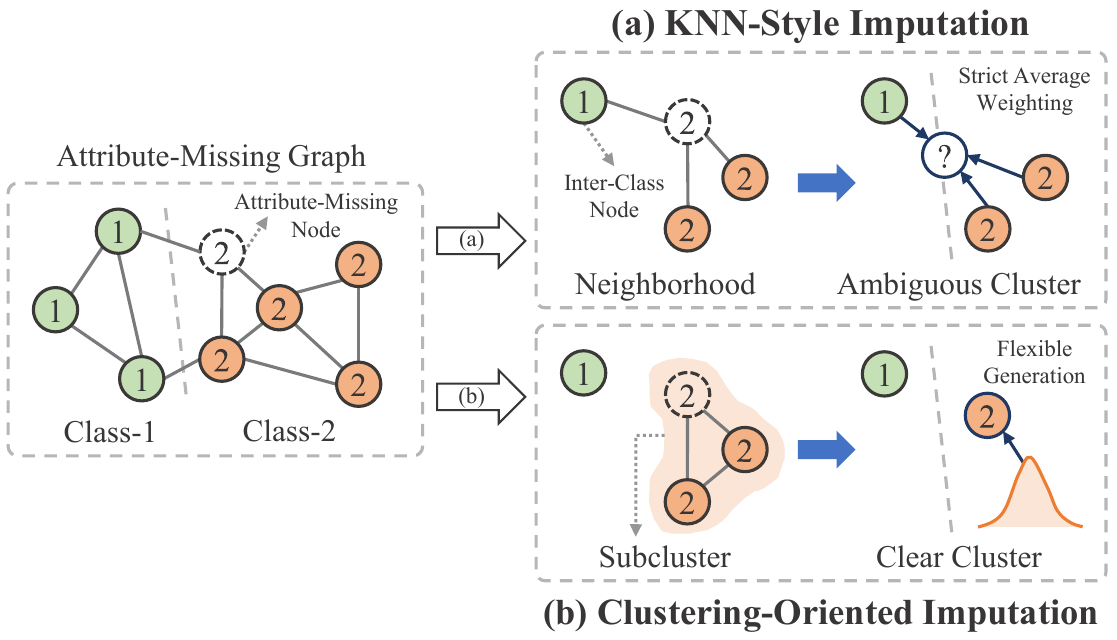}
	\end{subfigure}
	\caption{Challenge and opportunity of attribute graph imputation. (a) The neighborhood derived from KNN may incorporate inter-class nodes, thus the imputation result may deviate from the true class, resulting in an ambiguous cluster structure. (b) The subcluster provides a clustering-oriented distribution to guide flexible imputation, thus the imputation result remains compact with intra-class nodes, which facilitates a clear cluster structure.}
	\label{fig:motivation}
\end{figure}

Graph clustering aims to mine the internal groups within an attribute graph, and has emerged as a fundamental unsupervised task in many fields of multimedia, such as multimedia recommendation \cite{mu2022learning, yu2023multi, mao2024cluster}, multimedia retrieval \cite{lu2021graph, han2021fine, yu2024gsd}, and multimodal fusion \cite{liu2021two, li2023graphmft, lu2024bi}. Recently, Graph Convolutional Network (GCN) based deep graph clustering has presented remarkable preponderance, which benefits from the powerful neighbor aggregation mechanism \cite{GCN, wang2025locality, AMGC}. In general, neighboring nodes in a graph are more likely to be within the same cluster, so the neighbor aggregation mechanism of GCN can enhance the category semantic information of each node naturally for clustering improvement. 

Existing deep graph clustering mainly focuses on the complete graph data, which assumes that each node within a graph is accessible \cite{DAEGC, guo2022graph, DGCLUSTER, DCGL, liu2024reliable}. However, due to privacy setting, copyright protection, storage failure, etc., the attribute vectors of some nodes may be absent in the practical application. In other words, some specific sample rows in the attribute matrix are entirely missing, while the topological graph is intact. For example, in the Amazon co-purchase graph \cite{chen2020co}, each node represents a product, and the node attribute is the word bag of the consumers' feedback. Users may only provide comments for partial products due to privacy concerns. Since the conventional GCN-based deep graph clustering models lack  specific measures for the attribute-missing scenario, the learned graph embedding may be corrupted, resulting in sub-optimal clustering \cite{AMGC}. The above phenomenon surges to a new and challenging task, namely attribute-missing graph clustering. 

To handle attribute-missing graph data, an effective solution is to integrate the data completion theory into GCN-based graph learning \cite{SAT, SVGA, AMGC, RITR}. The relevant models can be roughly divided into two modules, including data imputation and imputation refinement (i.e., I-step and R-step). In I-step, existing methods project the nodes into a embedded space via GCN, and then estimate the attribute representations of missing nodes by the $k$-Nearest Neighbor (KNN) \cite{AMGC}, shared-latent space assumption \cite{SAT}, etc. In R-step, the imputation results are expected to reconstruct the initial graph structure, aiming to leverage the prior structural information to optimize the estimated representation. The two-step framework has shown tremendous potential for attribute-missing graph clustering, but there are still some issues.

\textbf{On the one hand}, existing node imputation schemes fail to fully leverage the class-specific characteristics, and may not be suitable for clustering. They often rely on a strict assumption, such as KNN-style completion \cite{AMGC} that the missing node can be recovered by the neighborhood's linear combination, and SAT \cite{SAT} that there is a shared-latent space for perfect alignment between the graph structure and attribute matrix. These assumptions restrict the search range of node imputation, which further decreases flexibility. Moreover, in the case of KNN, the local neighborhood based on topological graph may conflict with the class structure. As shown in Fig. \ref{fig:motivation}(a), the imputation node may deviate from the correct class distribution, and affects the clustering result adversely.

\textbf{On the other hand}, to optimize the learned embeddings, most methods use them to rebuild the initial graph directly, neglecting the intrinsic relationship between graphs and attributes. In the real-world graph data, there are usually only some specific attributes highly correlated to the presence of edges. For example, in the citation graph \cite{cora}, each node is digitized via the bag-of-words model, and each class represents the academic publications in a certain field. The words about research field are more critical to judging the citation structure, constituting the edge-wise attributes. Besides, the publications in different classes express the research field by diverse words, which means the intra/inter-class nodes tend to hold the consistent/distinguishable edge-wise attributes. 

\begin{figure*}[t]
	\centering
	\includegraphics[width=0.98\linewidth]{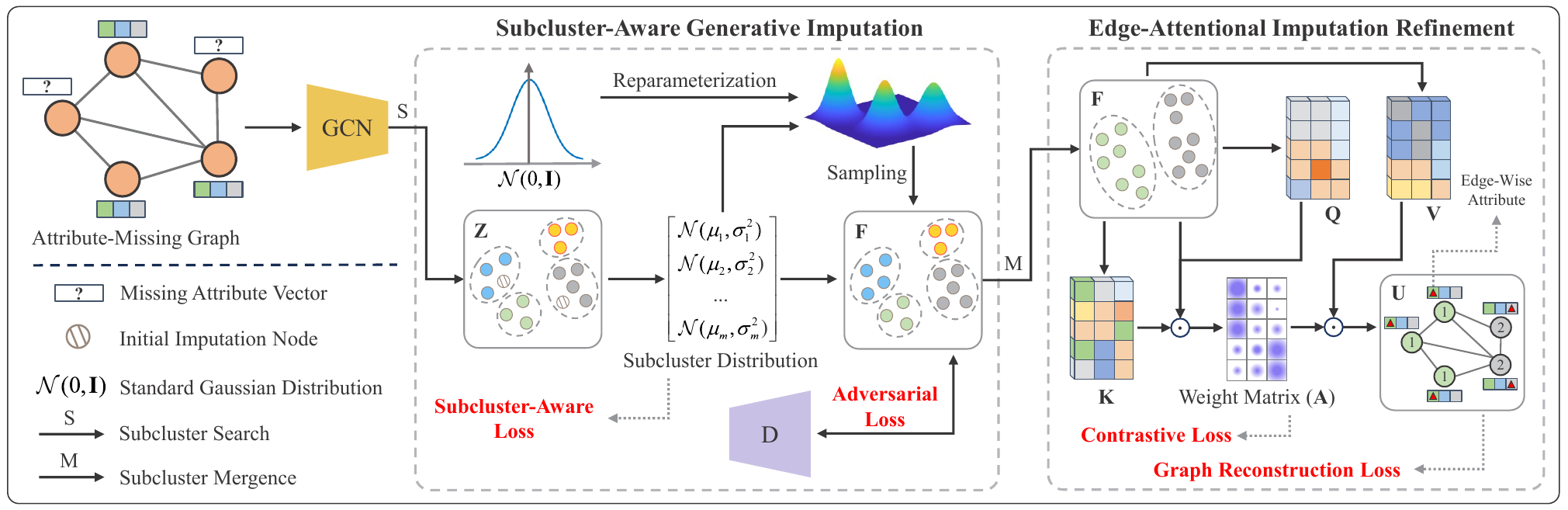}
	\caption{Overall flowchart of CGIR. Note that only one layer of edge attention networks is drawn. $\mathbf{Z}$ is the initial graph embedding, $\{\mathcal{N}(\mu_1, \sigma_1^2), \mathcal{N}(\mu_2, \sigma_2^2), ..., \mathcal{N}(\mu_m, \sigma_m^2) \}$ is a series of learned subcluster distributions, and $\mathbf{F}$ is the updated embedding by generative imputation. In edge attention networks, $\mathbf{Q}$, $\mathbf{K}$, and $\mathbf{V}$ is the query, key, and value embeddings respectively, $\mathbf{A}$ is the weight matrix, and $\mathbf{U}$ is the output embedding that emphasizes the edge-wise attributes. The framework is updated by alternate training shown in Section \ref{sec:training}.}
	\label{fig:overall}
\end{figure*}

In this paper, we propose a new attribute-missing graph clustering model termed Clustering-oriented Generative Imputation with reliable Refinement (CGIR). The overall pipeline is illustrated in Fig. \ref{fig:overall}, and the main contributions are listed as follows.
\begin{itemize}
	\item CGIR divides deep graph clustering into the search and mergence of subclusters to guide generative imputation and refinement respectively. Hence, data completion and refinement collaborate within a unified framework.
	
	\item The subcluster-aware generator and discriminator are established to conduct clustering-friendly imputation via adversarial games. The absent attributes are sampled from the compact subcluster distributions, which promotes the imputation nodes to maintain true class-specific characteristics.
	
	\item The edge attention network is proposed to perceive the edge-wise attributes adaptively, and a contrastive learning loss is designed to pursue edge-wise consistency of intra-class nodes. By emphasizing edge-wise attributes in graph reconstruction, the prior adjacent structure is exploited fully for reliable imputation refinement.
\end{itemize}

\section{Related Work}

\subsection{Deep Graph Clustering}

Recently, GCN-based deep graph clustering has drawn extensive attention, which leverages the powerful neighbor aggregation scheme of GCN to reveal the category pattern. Existing models can be roughly divided into two paradigms, including graph auto-encoder and contrastive graph clustering. 

For graph auto-encoder based methods, they first map the nodes into a latent space, and then rebuild the initial graph, so as to guarantee the structure relationship of learned embedding \cite{cui2020adaptive, pan2018adversarially}. A self-supervised cluster optimization scheme \cite{DEC, DAEGC, GDCL, guo2022graph, mrabah2022escaping} is widely used to facilitate a clustering-friendly graph embedding. Some methods \cite{SDCN, AGCCN, ding2023graph} parallel multi-layer perceptron and GCN to concurrently infer the feature- and structure-level embeddings, thus mining the original data adequately.

For contrastive graph clustering, they first utilize the graph augmentation technique \cite{you2020graph, CMVRL, GCA} to derive multiple views, and then conduct contrastive learning for discriminability improvement \cite{peng2023dual, DMAC}. Some approaches \cite{NCLA, zeng2024multi} advocate learnable augmentation to pursue views suitable for clustering. Moreover, some methods \cite{PGCL, SCAGC, HSAN} introduce pseudo-labels to guide the selection of positive and negative pairs, thus those nodes with similar category semantic information are pushed close by clustering-oriented contrastive learning, which accelerates a distinguishable cluster distribution.

The aforementioned models have achieved impressive performance, while all of them assume that the attribute vector of each node is accessible. However, such assumption may not hold in real scenarios, which poses a challenge to the current models. In this paper, the node imputation and refinement modules are integrated into a unified framework by searching and merging subclusters.

\subsection{Attribute-Missing Graph Learning}

Attribute-missing graph learning is an emerging research topic, aiming to handle the graph-type data with some nodes having no attribute records \cite{he2022analyzing}. The attribute-missing circumstance poses a huge challenge for graph embedding learning. Many studies attempt to join the imputation procedure into GCN-based deep model. SAT \cite{SAT} supposes that graph structure and attribute matrix can be projected into a shared-latent space. The available data is used to train a projector for node completion. SVGA \cite{SVGA} introduces the variational graph auto-encoder to estimate the data distribution, so as to sample missing nodes. RITR \cite{RITR} concatenates the attribute- and graph-level embeddings for node completion, and then optimizes the imputation result according to the affinity structure.

Most attribute-missing graph learning methods are established for the supervised task, and may be unsuitable for unsupervised graph clustering. AMGC \cite{AMGC} published in 2024 is described as the first attribute-missing graph clustering model, which utilizes the style of KNN for node imputation. In this paper, we plan to develop subcluster-aware generative imputation to endow the completion nodes with correct clustering semantic information, and design edge-attentional imputation refinement that promotes the imputation nodes to preserve the prior structure relationship.

\section{Methodology}

In this section, the process of CGIR is elaborated. As shown in Fig. \ref{fig:overall}, CGIR first learns the local subcluster distributions to generate missing nodes, and then perceives the edge-wise attributes to perform graph reconstruction for imputation refinement. 

\textbf{Notation.} The matrix and vector are written as uppercase and lowercase, respectively. Both $\mathbf{X}_i$ and $x_i$ denote the $i$-th row of $\mathbf{X}$. 

\subsection{Problem Definition} 

Define $\{\mathbf{X}, \mathbf{G}, \mathbf{h}\}$ as the attribute-missing graph data with $n$ nodes and $c$ classes. $\mathbf{X} \in \mathbb{R}^{n\times d}$ is the attribute matrix. $\mathbf{G} \in \mathbb{R}^{n\times n}$ is the adjacent graph. $\mathbf{h} \in \mathbb{R}^{n\times 1}$ is an indication vector of whether the node is available, that is, $h_i = 1$ if the $i$-th node is accessible, otherwise $h_i = 0$.

CGIR first utilizes a GCN to derive the graph embedding
\begin{equation}
	\label{eq:gcn_z}
	\begin{array}{c}
		\mathbf{Z} = \mathrm{GCN}(\mathbf{X}, \mathbf{G}).
	\end{array}
\end{equation}
The graph embeddings of missing nodes are filled by the available neighbors due to the neighbor aggregation mechanism of GCN, which can be seen as an initial imputation.

Based on the embedding $\mathbf{Z}$, we regard deep graph clustering as the search and mergence of subclusters, to guide the subcluster-aware generative imputation (i.e., I-step) and edge-attentional imputation refinement (i.e., R-step) respectively. In this way, node completion and refinement are integrated into a unified framework.

\subsection{Subcluster-Aware Generative Imputation} 

As discussed in Section \ref{sec:intro}, most imputation approaches are not tailored for clustering. Therefore, we use the multivariate Gaussian distribution to model the local subcluster, and sample the missing nodes from the underlying subcluster distribution to preserve the class semantic information. Meanwhile, we design the subcluster-aware loss to facilitate compact subclusters, and introduce the adversarial training mechanism to guarantee the sampling quality, further achieving clustering-oriented generative imputation.

\subsubsection{\textbf{Subcluster Distribution Learning}}

Compared with the holistic cluster, the more compact subcluster is employed to support the generative imputation, since it reflects the local distribution more accurately. Besides, different from the KNN-style neighborhood, the subcluster focuses on the class-relevant characteristics and tends to gather the intra-class nodes as displayed in Fig. \ref{fig:motivation}(b). In this part, we first compute the soft indicator between nodes and subclusters, and then convert it to multiple Gaussian subcluster distributions. Finally, we propose the subcluster-aware loss to constrain the volume of sampling space, leading to compact subclusters that exploit the class-specific characteristics more precisely.

Firstly, the graph embedding $\mathbf{Z}$ is divided into $m$ subclusters by agglomerative hierarchical clustering. Denote $\{\mathbf{U}_1, \mathbf{U}_2, ..., \mathbf{U}_m\}$ as the resultant sub-centroids (i.e., the center of subcluster). The element in soft assignment $\mathbf{P} \in \mathbb{R}^{n\times m}$ between nodes and subclusters can be calculated as
\begin{equation}
	\label{eq:similarity}
	\begin{array}{c}
		p_{ij} = 
		\frac{\left(1+||\mathbf{Z}_i - \mathbf{U}_j||_2^2\right)^{-1}}{\sum \limits_{b}^m \left(1+||\mathbf{Z}_i - \mathbf{U}_b||_2^2\right)^{-1}},
	\end{array}
\end{equation}
where $||\cdot||_2$ represents the $\ell_{2}$ norm, and $p_{ij}$ records the probability that the $i$-th node belongs to the $j$-th subcluster. 

Then, each column of $\mathbf{P}$ is regarded as the contributions of nodes to a subcluster. The multivariate Gaussian distribution is introduced to model the subcluster region. For the $j$-th subcluster distribution $\mathcal{N}(\mu_j, \sigma_j^2)$, the corresponding maximum likelihood estimations are
\begin{equation}
	\label{eq:distribution}
	\begin{array}{c}
		\mu_j = \frac{1}{\sum \limits_i^n p_{ij}} \sum \limits_i^n p_{ij}\mathbf{Z}_i, \\
		\sigma_j^2 = \frac{1}{\sum \limits_i^n p_{ij}} \sum \limits_i^n p_{ij}(\mathbf{Z}_i - \mu_j)^{\mathrm{T}}(\mathbf{Z}_i - \mu_j).
	\end{array}
\end{equation}

Finally, to pursue compact and high-quality subclusters, the sampling space revealed by the subcluster distribution should have a small volume. The data range of multivariate Gaussian distribution can be seen as a high-dimensional spheroid, and the eigenvalues of covariance matrix are the corresponding semi-axis lengths \cite{do2008multivariate}. Hence, the sampling space volume of subcluster distribution $\mathcal{N}(\mu_j, \sigma_j^2)$ can be quantified as
\begin{equation}
	\label{eq:volume}
	\begin{array}{c}
		\frac{\pi^{\widehat{d}/2}}{\Gamma\left(\frac{\widehat{d}}{2}+1\right)} \prod \limits_i^{\widehat{d}} \alpha_i^j,
	\end{array}
\end{equation}
where $\Gamma(\cdot)$ is the Gamma function, $\widehat{d}$ is the dimension of $\mathcal{N}(\mu_j, \sigma_j^2)$, and $\alpha_i^j$ is the $i$-th eigenvalue of $\sigma_j^2$. In model training, $\widehat{d}$ represents the neuron number of the output layer in GCN, and is a fixed constant. With reference to Eq. (\ref{eq:volume}), the subcluster-aware loss is devised as
\begin{equation}
	\label{eq:sub_loss}
	\begin{array}{c}
		{{\mathcal{L}}_{sub}} = \frac{1}{m} \sum \limits_i^m det\left(\sigma_i^{2} \right),
	\end{array}
\end{equation}
where $det(\cdot)$ means the determinant of a square matrix.

\subsubsection{\textbf{Generative Imputation and Discriminator}}

Based on the resultant $m$ subcluster distributions, the reparameterization trick is used to generate the imputation embedding. Furthermore, to evaluate and optimize the sampling result, a discriminator is designed to realize the adversarial training with the subcluster-aware generator. 

Since data sampling is a discrete process, the reparameterization rule \cite{trick} is adopted to avoid gradient interrupt. Supposing the $i$-th node is absent (i.e., $h_i=0$), the sampling procedure can be formulized as
\begin{equation}
	\label{eq:sample}
	\begin{array}{c}
		\epsilon_i \sim \mathcal{N}(0, \mathbf{I}), \\
		\widehat{\mathbf{Z}}_i = \sum \limits_j^{m} p_{ij}\left(\mu_j +  \epsilon_i \sigma_j\right),
	\end{array}
\end{equation}
where $\mathcal{N}(0, \mathbf{I})$ is the standard multivariate Gaussian distribution, and $\widehat{\mathbf{Z}}_i$ is the sampling result. With the rough imputation obtained by Eq. (\ref{eq:gcn_z}), the embedding is updated as
\begin{equation}
	\label{eq:F_calculation}
	\begin{array}{c}
		{{\mathbf{F}_i}}=\left\{\begin{array}{l}
			{\frac{1}{2}(\mathbf{Z}_i + \widehat{\mathbf{Z}}_i)}, h_i = 0,\\
			{\mathbf{Z}_i}, h_i = 1,
		\end{array} \right.
	\end{array}
\end{equation}
which means the generative imputation only applied to the missing nodes. $\mathbf{F}$ is the updated graph embedding, and the above procedure constitutes the subcluster-aware generator.

Then, we establish the subcluster discriminator $D(\cdot)$ to evaluate the sampling quality. Defining the $m$ subclusters as real classes, the discriminator $D(\cdot)$ performs multi-classification on $\mathbf{F}$, which can be formulized as 
\begin{equation}
	\label{eq:discriminator}
	\begin{array}{c}
		\mathbf{R} = D(\mathbf{F}),
	\end{array}
\end{equation}
where $\mathbf{R} \in \mathbb{R}^{n\times (m + 1)}$ is the probability matrix, and the last column in $\mathbf{R}$ records the resultant probabilities that nodes belong to the fake class. On this basis, we design the adversarial loss from two perspectives to guide the training of subcluster-aware generator and discriminator, respectively. From the perspective of subcluster-aware generator, the adversarial loss aims to divide each node into the correct subcluster, which can be expressed as
\begin{equation}
	\label{eq:gen}
	\begin{array}{c}
		{\mathcal{L}}_{ad1} = \frac{1}{n} \sum \limits_i^n CE(\widehat{\mathbf{P}}_i,\mathbf{R}_i),
	\end{array}
\end{equation}
where $\widehat{\mathbf{P}} \in \mathbb{R}^{n\times (m + 1)}$ is the extended $\mathbf{P}$ that the last column is all-zero, and $CE(\widehat{\mathbf{P}}_i,\mathbf{R}_i)$ treats $\widehat{\mathbf{P}}_i$ as the target distribution to calculate cross entropy. From the perspective of discriminator, all imputation nodes should be divided into the fake class, which minimizes
\begin{equation}
	\label{eq:dis}
	\begin{array}{c}
		{\mathcal{L}}_{ad2} = \frac{1}{n_0} \sum \limits_{h_i=0} CE(\vec{\mathbf{0}},\mathbf{R}_i) + \frac{1}{n_1} \sum \limits_{h_i=1} CE(\widehat{\mathbf{P}}_i,\mathbf{R}_i),
	\end{array}
\end{equation}
where $n_0$ and $n_1$ are the number of absent and available nodes respectively, and $\vec{\mathbf{0}} = (0, 0, ..., 1)$ is the ideal fake class indicator. By optimizing Eqs. (\ref{eq:gen}) and (\ref{eq:dis}) alternately, the mutual game \cite{adversarial} between subcluster-aware generator and discriminator is achieved to further pursue the real and clustering-oriented imputation results that align the correct subcluster distributions.

In summary, the proposed subcluster-aware generative imputation estimates the missing attributes based on the subcluster distributions, which reduces the risk of inter-class nodes involved in data completion, and promotes clustering-friendly imputation. In the next, we aim to further refine the imputation result via the prior graph structure.

\subsection{Edge-Attentional Imputation Refinement}

As described in Section \ref{sec:intro}, existing methods generally optimize the imputation nodes based on graph reconstruction, while neglects the edge-wise attributes that are closely related to the presence of edges. In this part, we propose the Edge Attention Network (EAN) to adaptively detect edge-wise attributes, and merge multiple subclusters to guide contrastive learning, which further accelerates the edge-wise distinction of nodes across different clusters. On this basis, the graph reconstruction is performed via the binary classification theory for reliable imputation refinement.

\subsubsection{\textbf{Edge Attention Network}}

To perceive the edge-wise attributes, each node should pay attention to the embeddings of its neighbors in the graph. Inspired by the self-attention mechanism \cite{attention}, we develop EAN to calculate the weight vector for each node according to the link relationship in the initial graph, such that the edge-wise attributes can be emphasized by feature weighting. Moreover, contrastive learning is introduced to enforce the intra-class nodes to hold consistent edge-wise attributes. The detailed process of EAN is as follows.

Firstly, the entire graph embedding matrix $\mathbf{F}$ is regarded as a sequence, and each node is set as a token. The query $\mathbf{Q}$, key $\mathbf{K}$, and value $\mathbf{V}$ are computed as
\begin{equation}
	\label{eq:EAN1}
	\begin{array}{c}
		\mathbf{Q} = \mathbf{F}\mathbf{W}_1, \mathbf{K} = \mathbf{F}\mathbf{W}_2, \mathbf{V} = \mathbf{F}\mathbf{W}_3,
	\end{array}
\end{equation}
where $\mathbf{W}_1$, $\mathbf{W}_2$, and $\mathbf{W}_3$ are the learnable parameter matrix, and have the identical dimension.

Then, for the $i$-th node, the corresponding neighbors in the initial graph $\mathbf{G}$ are adopted to derive the weight vector $\mathbf{A}_i$ as
\begin{equation}
	\label{eq:EAN2}
	\begin{array}{c}
		\mathbf{A}_{i}=softmax\Bigg(\frac{1}{|\xi(i)|}\sum \limits_{j\sim\xi(i)}\widehat{\mathbf{Q}}_{i}\odot\widehat{\mathbf{K}}_{j}\Bigg/_{\sqrt{d^{\prime}}}\Bigg),
	\end{array}
\end{equation}
where $\xi(i)$ is the neighbor set of the $i$-th node, $\odot$ represents the Hadamard product, $\widehat{\mathbf{Q}}_{i}$ and $\widehat{\mathbf{K}}_{j}$ denote the $\ell_{2}$ norm normalization results of $\mathbf{Q}_i$ and $\mathbf{K}_j$ respectively, and $d^{\prime}$ is the dimension value of $\mathbf{Q}_i$. By this manner, each node can perceive the attribute-level relation with neighbors, thus endowing the edge-wise attributes with larger weights. A series of weight vectors $\{\mathbf{A}_1,\mathbf{A}_2,...,\mathbf{A}_n\}$ are concatenated as rows to form the weight matrix $\mathbf{A}$.

In real-world applications, the nodes of different classes usually present diverse edge-wise attributes. In other words, the weight vectors of intra/inter-class nodes should be similar/distinct, which can be realized by contrastive learning
\begin{equation}
	\label{eq:contrastive_a}
	\begin{array}{c}
		\Omega(\mathbf{A}) = -\frac{1}{n}\sum \limits_i^n \left(log \frac{\sum \limits_{\mathbf{Y}_j=\mathbf{Y}_i} exp(\theta(\mathbf{A}_i, \mathbf{A}_j) / \tau)}{\sum \limits_{j}^{n} exp(\theta(\mathbf{A}_i, \mathbf{A}_j) / \tau)} \right), 
	\end{array}
\end{equation}
where $exp(\cdot)$ is the natural exponential function, $\theta(\cdot)$ means the cosine similarity, $\tau$ is the temperature parameter, $\mathbf{Y} \in \mathbb{R}^{n\times 1}$ is the pseudo-label sequence obtained by performing agglomerative hierarchical clustering on the subclusters, and $\mathbf{Y}_j = \mathbf{Y}_i$ indicates the $j$- and $i$-th nodes have the same cluster label. The above contrastive learning treats the intra/inter-class nodes as positive/negative pairs, such that the class-relevant information is incorporated into EAN to ameliorate the weight matrix $\mathbf{A}$, and further boost a clear cluster structure. 

Finally, the output $\mathbf{U}$ that highlights the edge-wise attributes is computed as
\begin{equation}
	\label{eq:EAN3}
	\begin{array}{c}
		\mathbf{U} = \mathbf{V}\odot \mathbf{A}.
	\end{array}
\end{equation}

As with other deep neural networks, multiple EANs can be arranged consecutively to mine multi-level edge-wise attributes. Based on Eq. (\ref{eq:contrastive_a}), the joint contrastive loss is
\begin{equation}
	\label{eq:contrastive_loss}
	\begin{array}{c}
		{{\mathcal{L}}_{con}} = \frac{1}{L} \sum \limits_i^L \Omega(\mathbf{A}^{(i)}),
	\end{array}
\end{equation}
where $L$ is the layer number of EANs, and $\mathbf{A}^{(i)}$ is the learnable weight matrix of the $i$-th layer EAN.

\subsubsection{\textbf{Graph Reconstruction via Binary Classification}}

Each edge weight in the initial graph $\mathbf{G}$ can be treated as the probability that two nodes are linked. Hence, the graph reconstruction is converted into $n^2$ binary classification problems. With the embedding $\mathbf{U}$, the simple but effective dot-product weighting \cite{DAEGC} is used to predict the structural graph
\begin{equation}
	\label{eq:S_calculation}
	\begin{array}{c}
		\mathbf{S} = sigmoid(\mathbf{U}\mathbf{U}^{\mathrm{T}}).
	\end{array}
\end{equation}

Referring to the binary cross entropy, the graph reconstruction loss is devised as
\begin{equation}
	\label{eq:gra_loss}
	\begin{array}{c}
		{{\mathcal{L}}_{gra}} = -\frac{1}{n^2} \sum \limits_{i,j}^{n}\left(G_{ij}logS_{ij} + (1-G_{ij})log(1-S_{ij}) \right).
	\end{array}
\end{equation}
Someone may be concerned about the applicability of Eq. (\ref{eq:gra_loss}) for a non-probabilistic graph where the edge weight in $\mathbf{G}$ may be greater than 1. In this circumstance, the initial graph $\mathbf{G}$ can be pre-processed by the Sigmoid function to calculate Eq. (\ref{eq:gra_loss}).

To sum up, the edge-attentional imputation refinement adjusts the learned embedding to emphasize the edge-wise attributes for more precise structure reconstruction, which exploits the given graph more adequately to drive reliable imputation refinement.

\subsection{Model Training}
\label{sec:training}

Since the proposed clustering-oriented imputation module involves the generative adversarial mechanism, we adopt the alternative training strategy. 

In each backpropagation, all network parameters except for the discriminator $D(\cdot)$ are frozen to minimize ${\mathcal{L}}_{ad2}$ shown in Eq. (\ref{eq:dis}). Then, only the parameters of discriminator $D(\cdot)$ are frozen to optimize the overall loss
\begin{equation}
	\label{eq:overall_loss1}
	\begin{array}{c}
		{\mathcal{L}} = {{\mathcal{L}}_{sub}} + \lambda_1 {{\mathcal{L}}_{ad1}} + \lambda_2  {{\mathcal{L}}_{con}} + \lambda_3 {{\mathcal{L}}_{gra}},
	\end{array}
\end{equation}
where $\lambda_1$, $\lambda_2$, and $\lambda_3$ are trade-off parameters. The above two steps constitute an epoch. Adam \cite{Adam} is used to train CGIR.

\section{Complexity Analysis}

In this section, the averaged time complexity of each module in CGIR is provided in detail. To evade excessive symbol definitions and improve readability, we only consider the influence of node size $n$ and subcluster number $m$.

In each forward propagation, the computation complexity of graph embedding learning shown in Eq. (\ref{eq:gcn_z}) is $O(n^2)$ to support the neighbor aggregation scheme. Then, agglomerative hierarchical clustering requires $O(n^2-nm)$ to derive subclusters, and Eq. (\ref{eq:similarity}) needs $O(nm)$ to calculate the soft assignment matrix between nodes and subclusters. The time complexity of Eq. (\ref{eq:distribution}) is also $O(nm)$ to model the Gaussian subcluter distributions. Eq. (\ref{eq:sample}) and Eq. (\ref{eq:F_calculation}) consume $O(n)$ to conduct clustering-oriented generative imputation. After that, the edge attention network spends $O(n)$ to perceive the edge-wise attributes. Finally, Eq. (\ref{eq:S_calculation}) requires $O(n^2)$ to recover the initial graph structure.

In each back propagation, the subcluster-aware loss ${{\mathcal{L}}_{sub}}$ needs $O(m)$ to promote compact subcluster distributions. The time complexity of both adversarial losses ${{\mathcal{L}}_{ad1}}$ and ${{\mathcal{L}}_{ad2}}$ is $O(n\left(m+1\right))$ to guarantee the quality of generative imputation. The contrastive learning loss ${{\mathcal{L}}_{con}}$ consumes $O(n)$ to pursue the consistent edge-wise attributes among intra-class nodes. ${\mathcal{L}}_{gra}$ needs $O(n^2)$ to perform graph reconstruction via binary classification for reliable imputation refinement.

Overall, the time complexity of CGIR is $O(n^2+nm)$. In general, the number of subclusters is much smaller than that of samples (i.e., $m \ll n$), so the averaged computation complexity of CGIR can be seen as $O(n^2)$, which is moderate compared to advanced attribute-missing graph learning models \cite{RITR, SAT, SVGA}. The experimental verification can be found in Section \ref{sec:comparison}.

\section{Experiments}

\subsection{Benchmark Datasets}

\begin{table}[t]
	\tabcolsep=0.12cm
	\centering
	\small
	\caption{Details of benchmark datasets.}
	\begin{tabular}{ccccc}
		\toprule
		\textbf{Dataset} & \textbf{Samples} & \textbf{Edges} & \textbf{Features} & \textbf{Classes} \\ 
		\midrule
		\textbf{UAT} & 1190 & 13599 & 239 & 4 \\
		\textbf{Cora}	& 2708 & 5429 & 1433 & 7 \\
		\textbf{AMAP} 	& 7650 & 119081 & 745 & 8 \\
		\textbf{PUBMED}	& 19717 & 44324 & 500 & 3 \\
		\bottomrule
	\end{tabular}
	\label{tab:datasets}
\end{table}

\begin{table*}[t]
	\tabcolsep=0.1cm
	\small
	\centering
	\caption{Average clustering performance of several methods on graph data with different missing ratios. Best values are decorated with bold. 'OOM' denotes out of memory on RTX-3090 24GB. }
	\begin{tabular}{cccccc|cccc|cccc|cccc}
		\toprule
		\multirow{2.5}{*}{\makecell[c]{Missing \\ Ratio}} & \makecell[c]{\multirow{2.5}{*}{Method}} & \multicolumn{4}{c}{UAT} & \multicolumn{4}{c}{Cora} & \multicolumn{4}{c}{AMAP} & \multicolumn{4}{c}{PUBMED}\\
		\cmidrule{3-18} 
		\makecell[c]{\multirow{2.5}{*}{}} & \makecell[c]{\multirow{2.5}{*}{}} & \makecell[c]{ACC} & \makecell[c]{NMI} & \makecell[c]{ARI} & \makecell[c]{F1} & \makecell[c]{ACC} & \makecell[c]{NMI} & \makecell[c]{ARI} & \makecell[c]{F1} & \makecell[c]{ACC} & \makecell[c]{NMI} & \makecell[c]{ARI} & \makecell[c]{F1} & \makecell[c]{ACC} & \makecell[c]{NMI} & \makecell[c]{ARI} & \makecell[c]{F1}\\
		\midrule
		\makecell[c]{\multirow{11}{*}{0.2}} & \makecell[c]{DAEGC} & \makecell[c]{50.38} & \makecell[c]{18.14} & \makecell[c]{17.46} & \makecell[c]{48.05} & \makecell[c]{65.96} & \makecell[c]{47.73} & \makecell[c]{43.31} & \makecell[c]{60.77} & \makecell[c]{71.35} & \makecell[c]{62.73} & \makecell[c]{53.54} & \makecell[c]{67.54} & \makecell[c]{OOM} & \makecell[c]{OOM} & \makecell[c]{OOM} & \makecell[c]{OOM}\\
		\makecell[c]{\multirow{11}{*}{}} & \makecell[c]{SDCN} & \makecell[c]{44.95} & \makecell[c]{12.82} & \makecell[c]{11.40} & \makecell[c]{43.13} & \makecell[c]{30.61} & \makecell[c]{14.38} & \makecell[c]{7.79} & \makecell[c]{28.47} & \makecell[c]{40.02} & \makecell[c]{24.89} & \makecell[c]{13.35} & \makecell[c]{36.45} & \makecell[c]{44.55} & \makecell[c]{5.90} & \makecell[c]{3.66} & \makecell[c]{42.51}\\
		\makecell[c]{\multirow{11}{*}{}} & \makecell[c]{AGC-DRR} & \makecell[c]{44.96} & \makecell[c]{11.31} & \makecell[c]{10.41} & \makecell[c]{39.69} & \makecell[c]{40.40} & \makecell[c]{19.62} & \makecell[c]{12.53} & \makecell[c]{32.40} & \makecell[c]{55.35} & \makecell[c]{43.72} & \makecell[c]{34.89} & \makecell[c]{44.13} & \makecell[c]{52.49} & \makecell[c]{7.45} & \makecell[c]{9.41} & \makecell[c]{47.55}\\
		\makecell[c]{\multirow{11}{*}{}} & \makecell[c]{SAT} & \makecell[c]{43.70} & \makecell[c]{17.18} & \makecell[c]{11.30} & \makecell[c]{41.65} & \makecell[c]{45.13} & \makecell[c]{27.09} & \makecell[c]{17.64} & \makecell[c]{40.74} & \makecell[c]{21.12} & \makecell[c]{2.93} & \makecell[c]{1.45} & \makecell[c]{14.87} & \makecell[c]{61.31} & \makecell[c]{23.57} & \makecell[c]{24.94} & \makecell[c]{62.83}\\
		\makecell[c]{\multirow{11}{*}{}} & \makecell[c]{SVGA} & \makecell[c]{46.72} & \makecell[c]{21.01} & \makecell[c]{14.56} & \makecell[c]{43.67} & \makecell[c]{55.65} & \makecell[c]{43.21} & \makecell[c]{35.89} & \makecell[c]{48.23} & \makecell[c]{42.85} & \makecell[c]{28.51} & \makecell[c]{15.79} & \makecell[c]{37.29} & \makecell[c]{40.64} & \makecell[c]{0.73} & \makecell[c]{0.64} & \makecell[c]{28.33}\\
		\makecell[c]{\multirow{11}{*}{}} & \makecell[c]{HSAN} & \makecell[c]{48.14} & \makecell[c]{20.14} & \makecell[c]{13.75} & \makecell[c]{47.72} & \makecell[c]{63.31} & \makecell[c]{46.91} & \makecell[c]{38.73} & \makecell[c]{57.92} & \makecell[c]{75.97} & \makecell[c]{65.74} & \makecell[c]{56.47} & \makecell[c]{70.82} & \makecell[c]{OOM} & \makecell[c]{OOM} & \makecell[c]{OOM} & \makecell[c]{OOM}\\
		\makecell[c]{\multirow{11}{*}{}} & \makecell[c]{COMPLETER} & \makecell[c]{47.08} & \makecell[c]{18.46} & \makecell[c]{16.45} & \makecell[c]{46.25} & \makecell[c]{30.64} & \makecell[c]{2.57} & \makecell[c]{1.18} & \makecell[c]{15.10} & \makecell[c]{30.62} & \makecell[c]{16.87} & \makecell[c]{6.33} & \makecell[c]{25.40} & \makecell[c]{51.67} & \makecell[c]{13.12} & \makecell[c]{9.90} & \makecell[c]{47.60}\\
		\makecell[c]{\multirow{11}{*}{}} & \makecell[c]{APADC} & \makecell[c]{45.56} & \makecell[c]{19.08} & \makecell[c]{14.69} & \makecell[c]{45.13} & \makecell[c]{30.47} & \makecell[c]{2.73} & \makecell[c]{0.73} & \makecell[c]{13.39} & \makecell[c]{31.76} & \makecell[c]{15.55} & \makecell[c]{6.17} & \makecell[c]{20.50} & \makecell[c]{56.28} & \makecell[c]{20.59} & \makecell[c]{24.51} & \makecell[c]{58.23}\\
		\makecell[c]{\multirow{11}{*}{}} & \makecell[c]{GIMVC} & \makecell[c]{41.62} & \makecell[c]{19.70} & \makecell[c]{12.85} & \makecell[c]{38.93} & \makecell[c]{37.79} & \makecell[c]{18.05} & \makecell[c]{9.09} & \makecell[c]{37.40} & \makecell[c]{39.28} & \makecell[c]{23.14} & \makecell[c]{8.37} & \makecell[c]{31.62} & \makecell[c]{52.35} & \makecell[c]{15.50} & \makecell[c]{13.36} & \makecell[c]{52.18}\\
		\makecell[c]{\multirow{11}{*}{}} & \makecell[c]{CGIR (ours)} & \makecell[c]{\textbf{52.03}} & \makecell[c]{\textbf{21.09}} & \makecell[c]{\textbf{19.39}} & \makecell[c]{\textbf{50.55}} & \makecell[c]{\textbf{68.58}} & \makecell[c]{\textbf{49.38}} & \makecell[c]{\textbf{44.69}} & \makecell[c]{\textbf{66.65}} & \makecell[c]{\textbf{76.93}} & \makecell[c]{\textbf{67.16}} & \makecell[c]{\textbf{57.80}} & \makecell[c]{\textbf{72.93}} & \makecell[c]{\textbf{67.54}} & \makecell[c]{\textbf{30.68}} & \makecell[c]{\textbf{29.40}} & \makecell[c]{\textbf{67.29}}\\
		\midrule
		\makecell[c]{\multirow{11}{*}{0.4}} & \makecell[c]{DAEGC}& \makecell[c]{47.90} & \makecell[c]{16.22} & \makecell[c]{15.27} & \makecell[c]{45.33} & \makecell[c]{62.81} & \makecell[c]{43.43} & \makecell[c]{38.78} & \makecell[c]{60.05} & \makecell[c]{69.42} & \makecell[c]{61.89} & \makecell[c]{52.10} & \makecell[c]{66.43} & \makecell[c]{OOM} & \makecell[c]{OOM} & \makecell[c]{OOM} & \makecell[c]{OOM}\\
		\makecell[c]{\multirow{11}{*}{}} & \makecell[c]{SDCN} & \makecell[c]{38.12} & \makecell[c]{7.49} & \makecell[c]{5.00} & \makecell[c]{35.15} & \makecell[c]{31.53} & \makecell[c]{14.75} & \makecell[c]{5.56} & \makecell[c]{28.96} & \makecell[c]{30.51} & \makecell[c]{19.07} & \makecell[c]{7.41} & \makecell[c]{25.26} & \makecell[c]{39.78} & \makecell[c]{0.01} & \makecell[c]{0.03} & \makecell[c]{29.29}\\
		\makecell[c]{\multirow{11}{*}{}} & \makecell[c]{AGC-DRR} & \makecell[c]{43.59} & \makecell[c]{12.46} & \makecell[c]{11.47} & \makecell[c]{39.81} & \makecell[c]{38.85} & \makecell[c]{17.64} & \makecell[c]{11.53} & \makecell[c]{35.65} & \makecell[c]{51.20} & \makecell[c]{38.64} & \makecell[c]{30.95} & \makecell[c]{39.48} & \makecell[c]{51.83} & \makecell[c]{6.55} & \makecell[c]{8.51} & \makecell[c]{46.05}\\
		\makecell[c]{\multirow{11}{*}{}} & \makecell[c]{SAT} & \makecell[c]{44.29} & \makecell[c]{18.01} & \makecell[c]{11.74} & \makecell[c]{42.68} & \makecell[c]{35.12} & \makecell[c]{19.60} & \makecell[c]{11.23} & \makecell[c]{27.42} & \makecell[c]{20.26} & \makecell[c]{2.42} & \makecell[c]{10.07} & \makecell[c]{15.01} & \makecell[c]{57.64} & \makecell[c]{19.41} & \makecell[c]{22.76} & \makecell[c]{60.95}\\
		\makecell[c]{\multirow{11}{*}{}} & \makecell[c]{SVGA} & \makecell[c]{47.23} & \makecell[c]{19.03} & \makecell[c]{12.47} & \makecell[c]{46.15} & \makecell[c]{42.98} & \makecell[c]{27.88} & \makecell[c]{21.91} & \makecell[c]{37.17} & \makecell[c]{36.17} & \makecell[c]{20.07} & \makecell[c]{10.74} & \makecell[c]{30.81} & \makecell[c]{40.51} & \makecell[c]{0.63} & \makecell[c]{0.66} & \makecell[c]{28.29}\\
		\makecell[c]{\multirow{11}{*}{}} & \makecell[c]{HSAN} & \makecell[c]{44.40} & \makecell[c]{18.00} & \makecell[c]{11.43} & \makecell[c]{43.38} & \makecell[c]{61.04} & \makecell[c]{44.01} & \makecell[c]{36.56} & \makecell[c]{53.18} & \makecell[c]{69.42} & \makecell[c]{59.92} & \makecell[c]{50.11} & \makecell[c]{61.99} & \makecell[c]{OOM} & \makecell[c]{OOM} & \makecell[c]{OOM} & \makecell[c]{OOM}\\
		\makecell[c]{\multirow{11}{*}{}} & \makecell[c]{COMPLETER} & \makecell[c]{41.63} & \makecell[c]{17.60} & \makecell[c]{10.05} & \makecell[c]{38.39} & \makecell[c]{30.49} & \makecell[c]{2.01} & \makecell[c]{0.69} & \makecell[c]{14.13} & \makecell[c]{26.91} & \makecell[c]{9.23} & \makecell[c]{2.15} & \makecell[c]{19.78} & \makecell[c]{46.79} & \makecell[c]{4.72} & \makecell[c]{3.57} & \makecell[c]{40.17}\\
		\makecell[c]{\multirow{11}{*}{}} & \makecell[c]{APADC} & \makecell[c]{41.70} & \makecell[c]{15.46} & \makecell[c]{7.79} & \makecell[c]{40.56} & \makecell[c]{30.36} & \makecell[c]{2.06} & \makecell[c]{0.73} & \makecell[c]{13.84} & \makecell[c]{28.51} & \makecell[c]{9.27} & \makecell[c]{1.96} & \makecell[c]{17.71} & \makecell[c]{54.74} & \makecell[c]{15.41} & \makecell[c]{22.99} & \makecell[c]{49.35}\\
		\makecell[c]{\multirow{11}{*}{}} & \makecell[c]{GIMVC} & \makecell[c]{39.38} & \makecell[c]{18.14} & \makecell[c]{8.51} & \makecell[c]{34.03} & \makecell[c]{29.97} & \makecell[c]{8.32} & \makecell[c]{3.84} & \makecell[c]{25.60} & \makecell[c]{35.39} & \makecell[c]{17.77} & \makecell[c]{5.10} & \makecell[c]{26.47} & \makecell[c]{45.12} & \makecell[c]{7.16} & \makecell[c]{6.12} & \makecell[c]{45.40}\\
		\makecell[c]{\multirow{11}{*}{}} & \makecell[c]{CGIR (ours)} & \makecell[c]{\textbf{49.00}} & \makecell[c]{18.39} & \makecell[c]{\textbf{15.88}} & \makecell[c]{\textbf{46.56}} & \makecell[c]{\textbf{63.82}} & \makecell[c]{\textbf{44.94}} & \makecell[c]{\textbf{38.79}} & \makecell[c]{\textbf{62.67}} & \makecell[c]{\textbf{76.15}} & \makecell[c]{\textbf{65.58}} & \makecell[c]{\textbf{55.77}} & \makecell[c]{\textbf{71.04}} & \makecell[c]{\textbf{65.42}} & \makecell[c]{\textbf{28.35}} & \makecell[c]{\textbf{26.60}} & \makecell[c]{\textbf{64.14}}\\
		\midrule
		\makecell[c]{\multirow{11}{*}{0.6}} & \makecell[c]{DAEGC} & \makecell[c]{44.64} & \makecell[c]{14.44} & \makecell[c]{12.42} & \makecell[c]{40.29} & \makecell[c]{58.04} & \makecell[c]{39.16} & \makecell[c]{32.87} & \makecell[c]{55.42} & \makecell[c]{63.00} & \makecell[c]{54.56} & \makecell[c]{41.19} & \makecell[c]{54.21} & \makecell[c]{OOM} & \makecell[c]{OOM} & \makecell[c]{OOM} & \makecell[c]{OOM}\\
		\makecell[c]{\multirow{11}{*}{}} & \makecell[c]{SDCN} & \makecell[c]{31.11} & \makecell[c]{5.20} & \makecell[c]{1.58} & \makecell[c]{27.82} & \makecell[c]{27.11} & \makecell[c]{5.92} & \makecell[c]{0.70} & \makecell[c]{16.90} & \makecell[c]{30.12} & \makecell[c]{12.66} & \makecell[c]{3.14} & \makecell[c]{20.17} & \makecell[c]{39.75} & \makecell[c]{0.01} & \makecell[c]{0.01} & \makecell[c]{29.26}\\
		\makecell[c]{\multirow{11}{*}{}} & \makecell[c]{AGC-DRR} & \makecell[c]{42.95} & \makecell[c]{12.34} & \makecell[c]{10.51} & \makecell[c]{37.79} & \makecell[c]{34.49} & \makecell[c]{13.22} & \makecell[c]{9.08} & \makecell[c]{30.26} & \makecell[c]{53.73} & \makecell[c]{37.07} & \makecell[c]{30.13} & \makecell[c]{39.70} & \makecell[c]{50.68} & \makecell[c]{6.98} & \makecell[c]{7.89} & \makecell[c]{45.83}\\
		\makecell[c]{\multirow{11}{*}{}} & \makecell[c]{SAT} & \makecell[c]{39.33} & \makecell[c]{12.45} & \makecell[c]{6.88} & \makecell[c]{39.77} & \makecell[c]{32.16} & \makecell[c]{11.49} & \makecell[c]{5.09} & \makecell[c]{27.48} & \makecell[c]{20.43} & \makecell[c]{2.03} & \makecell[c]{0.88} & \makecell[c]{14.43} & \makecell[c]{53.71} & \makecell[c]{13.16} & \makecell[c]{21.69} & \makecell[c]{50.45}\\
		\makecell[c]{\multirow{11}{*}{}} & \makecell[c]{SVGA} & \makecell[c]{40.76} & \makecell[c]{13.74} & \makecell[c]{8.14} & \makecell[c]{34.76} & \makecell[c]{21.86} & \makecell[c]{2.02} & \makecell[c]{0.65} & \makecell[c]{17.32} & \makecell[c]{20.88} & \makecell[c]{2.54} & \makecell[c]{1.15} & \makecell[c]{14.89} & \makecell[c]{40.15} & \makecell[c]{0.26} & \makecell[c]{0.29} & \makecell[c]{30.08}\\
		\makecell[c]{\multirow{11}{*}{}} & \makecell[c]{HSAN} & \makecell[c]{41.27} & \makecell[c]{13.46} & \makecell[c]{9.78} & \makecell[c]{38.57} & \makecell[c]{53.04} & \makecell[c]{37.48} & \makecell[c]{28.10} & \makecell[c]{51.21} & \makecell[c]{62.61} & \makecell[c]{53.06} & \makecell[c]{42.14} & \makecell[c]{56.97} & \makecell[c]{OOM} & \makecell[c]{OOM} & \makecell[c]{OOM} & \makecell[c]{OOM}\\
		\makecell[c]{\multirow{11}{*}{}} & \makecell[c]{COMPLETER} & \makecell[c]{36.24} & \makecell[c]{11.39} & \makecell[c]{5.15} & \makecell[c]{32.60} & \makecell[c]{30.38} & \makecell[c]{1.70} & \makecell[c]{0.38} & \makecell[c]{12.42} & \makecell[c]{26.85} & \makecell[c]{4.68} & \makecell[c]{1.21} & \makecell[c]{15.62} & \makecell[c]{42.16} & \makecell[c]{1.69} & \makecell[c]{0.58} & \makecell[c]{34.16}\\
		\makecell[c]{\multirow{11}{*}{}} & \makecell[c]{APADC} & \makecell[c]{33.83} & \makecell[c]{9.17} & \makecell[c]{2.63} & \makecell[c]{33.40} & \makecell[c]{30.13} & \makecell[c]{1.78} & \makecell[c]{0.62} & \makecell[c]{14.06} & \makecell[c]{27.47} & \makecell[c]{5.96} & \makecell[c]{0.94} & \makecell[c]{16.14} & \makecell[c]{51.93} & \makecell[c]{12.47} & \makecell[c]{15.14} & \makecell[c]{47.26}\\
		\makecell[c]{\multirow{11}{*}{}} & \makecell[c]{GIMVC} & \makecell[c]{35.16} & \makecell[c]{13.70} & \makecell[c]{4.28} & \makecell[c]{29.16} & \makecell[c]{30.14} & \makecell[c]{5.08} & \makecell[c]{1.85} & \makecell[c]{22.00} & \makecell[c]{33.07} & \makecell[c]{12.71} & \makecell[c]{2.77} & \makecell[c]{22.45} & \makecell[c]{42.67} & \makecell[c]{4.35} & \makecell[c]{1.67} & \makecell[c]{40.06}\\
		\makecell[c]{\multirow{11}{*}{}} & \makecell[c]{CGIR (ours)} & \makecell[c]{\textbf{45.42}} & \makecell[c]{\textbf{15.09}} & \makecell[c]{\textbf{12.48}} & \makecell[c]{\textbf{41.84}} & \makecell[c]{\textbf{59.70}} & \makecell[c]{\textbf{40.10}} & \makecell[c]{\textbf{33.12}} & \makecell[c]{\textbf{57.26}} & \makecell[c]{\textbf{74.78}} & \makecell[c]{\textbf{64.32}} & \makecell[c]{\textbf{53.88}} & \makecell[c]{\textbf{69.93}} & \makecell[c]{\textbf{64.71}} & \makecell[c]{\textbf{26.79}} & \makecell[c]{\textbf{25.53}} & \makecell[c]{\textbf{64.38}}\\
		\bottomrule
	\end{tabular}
	\label{tab:comparsion}
\end{table*}

Four datasets are collected as benchmarks, including UAT \cite{SCAGC}, Cora \cite{cora}, AMAP \cite{amap}, and PUBMED \cite{pubmed}. Among them, UAT records the passenger's airport activity, Cora is a classical citation network, AMAP stores the co-purchase relationship between products from Amazon, and PUBMED is a citation network of biomedical publications. Table \ref{tab:datasets} shows the basic information of benchmark datasets.

\subsection{Evaluation Metrics}

Four metrics are adopted to comprehensively quantify the clustering result, including Accuracy (ACC), Normalized Mutual Information (NMI), Adjusted Rand Index (ARI), and F1-score (F1). The mathematical expression of metrics can be found in \cite{metric1, metric2}. The larger the metric value, the better the clustering performance. 

\subsection{Competitors and Settings}

\label{sec:setting}

We select four advanced graph clustering models (i.e., DAEGC \cite{DAEGC}, SDCN \cite{SDCN}, AGC-DRR \cite{AGC-DRR}, and HSAN \cite{HSAN}), and two attribute imputation networks (i.e., SAT \cite{SAT}, and SVGA \cite{SVGA}). Since there is few research for attribute-missing graph clustering, we regard the attribute matrix and topology graph as two views, and adopt three incomplete multi-view clustering methods as competitors, including COMPLETER \cite{COMPLETER}, APADC \cite{APADC}, and GIMVC \cite{GIMVC}.

The hyper-parameters of comparative algorithms are selected according to the original articles. For the proposed CGIR, the maximum epochs are 100. The number of sub-clusters $m$ is fixed as $c\times4$, where $c$ is the number of true classes. The temperature parameter $\tau$ in Eq. (\ref{eq:contrastive_a}) is fixed as 0.1. The trade-off parameters in Eq. (\ref{eq:overall_loss1}) are fixed as $\lambda_1=\lambda_2=\lambda_3=10$. For fairness, the final clustering result is derived by performing $k$-means \cite{kmeans} on the converged $\mathbf{F}$. All deep models are trained with an NVIDIA RTX-3090 GPU. Each method is repeated 10 times to report the average performance.

\subsection{Comparison Results}
\label{sec:comparison}

To simulate the real attribute-missing graph dataset, we randomly remove some rows in the attribute matrix with a predefined missing ratio, and hold the complete graph structure. In the attribute matrix, the rows corresponding to the missing nodes are initialized with all-zero vectors, which is consistent with the settings of competitors for a fair comparison. Table \ref{tab:comparsion} displays the average clustering performance of each algorithm with different missing ratios. Through observation and comparison, we summarize the following viewpoints. 
Firstly, CGIR presents the most outstanding capacity on clustering attribute-missing graphs. Benefiting from the proposed generative imputation and edge-attentional refinement, CGIR can estimate the missing nodes suitable for clustering, and further boost the imputation result based on the given structure relationship. The adverse impact of missing attributes is inhibited effectively to facilitate graph clustering. Secondly, The clustering ability of HSAN and DAEGC is relatively prominent over other comparative algorithms. Nevertheless, compared with the proposed CGIR, the performance of HSAN and DAEGC deteriorates obviously with the increase of missing ratios, especially on AMAP. The comparison further demonstrates the practicability of CGIR on processing attribute-missing graphs. Thirdly, The performance scores of incomplete multi-view clustering are inferior. These methods treat the topological graph as a feature view straightforwardly, and fail to exploit the structure-level semantic information sufficiently. 

\begin{figure}[t]
	\centering
	\begin{subfigure}{0.46\linewidth}
		\includegraphics[width=1\linewidth]{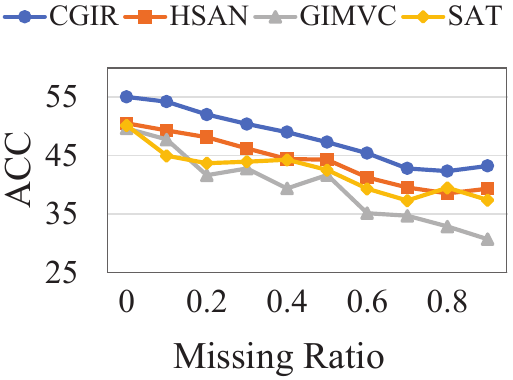}
		\caption{UAT}
	\end{subfigure}
	\hspace{0.02\linewidth}
	\begin{subfigure}{0.46\linewidth}
		\includegraphics[width=1\linewidth]{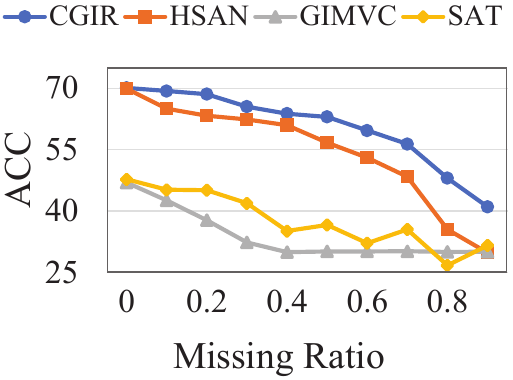}
		\caption{Cora}
	\end{subfigure}
	\caption{Comparison results on two benchmarks with extensive missing ratios (i.e., 0 to 0.9 with 0.1 interval).}
	\label{fig:missing}
\end{figure}

To further validate the advantage of CGIR, we tune the missing ratio from 0 to 0.9 for more sufficient comparison. The clustering results are exhibited in Fig. \ref{fig:missing}. The performance deteriorates gradually with the increase of missing ratio, which reveals the enormous adverse effects of missing attributes. The proposed CGIR still retains the optimal clustering ability even in the circumstance of a large number of missing nodes. Especially on Cora, the clustering accuracy of CGIR with 0.9 missing ratio surpasses that of GIMVC with 0.2 absent attributes. It is indicated that CGIR can derive subcluster distributions based on extremely scarce data to support imputation. 

\begin{figure}[t]
	\centering
	\begin{subfigure}{0.9\linewidth}
		\includegraphics[width=1\linewidth]{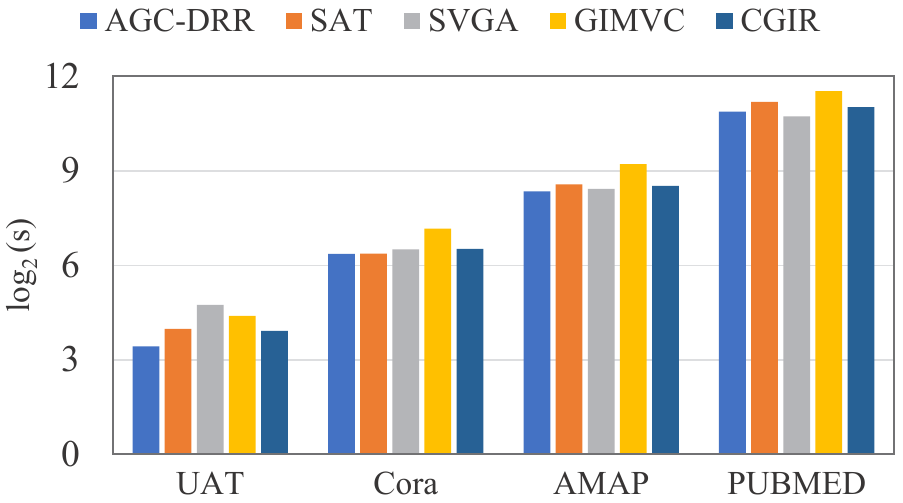}
	\end{subfigure}
	\caption{Running time (s) of graph learning models on four benchmarks with 0.2 missing ratio.}
	\label{fig:speed}
\end{figure}

Furthermore, the runtime of several deep graph learning models is compared. Fig. \ref{fig:speed} exhibits the efficiency comparison on benchmarks with 0.2 missing ratio. CGIR presents a moderate computational speed compared to advanced graph clustering competitors, which further demonstrates the practicability of new model. 

\subsection{Ablation Study}

\begin{table*}[t]
	\tabcolsep=0.1cm
	\centering
	\small
	\caption{Ablation comparison of main modules. Optimal metric scores are highlighted in bold.}
	\begin{tabular}{cccccc|cccc|cccc|cccc}
		\toprule
		\multirow{2.5}{*}{\makecell[c]{Missing \\ Ratio}} & \makecell[c]{\multirow{2.5}{*}{Variant}} & \multicolumn{4}{c}{UAT} & \multicolumn{4}{c}{Cora} & \multicolumn{4}{c}{AMAP} & \multicolumn{4}{c}{PUBMED}\\
		\cmidrule{3-18} 
		\makecell[c]{\multirow{2.5}{*}{}} & \makecell[c]{\multirow{2.5}{*}{}} & \makecell[c]{ACC} & \makecell[c]{NMI} & \makecell[c]{ARI} & \makecell[c]{F1} & \makecell[c]{ACC} & \makecell[c]{NMI} & \makecell[c]{ARI} & \makecell[c]{F1} & \makecell[c]{ACC} & \makecell[c]{NMI} & \makecell[c]{ARI} & \makecell[c]{F1} & \makecell[c]{ACC} & \makecell[c]{NMI} & \makecell[c]{ARI} & \makecell[c]{F1}\\
		\midrule
		\makecell[c]{\multirow{3}{*}{0.2}} & \makecell[c]{wo/GI} & \makecell[c]{49.78} & \makecell[c]{18.61} & \makecell[c]{17.10} & \makecell[c]{47.77} & \makecell[c]{64.65} & \makecell[c]{47.09} & \makecell[c]{41.38} & \makecell[c]{62.89} & \makecell[c]{74.37} & \makecell[c]{62.68} & \makecell[c]{51.60} & \makecell[c]{67.55} & \makecell[c]{65.91} & \makecell[c]{28.18} & \makecell[c]{26.91} & \makecell[c]{65.54}\\
		\makecell[c]{\multirow{3}{*}{}} & \makecell[c]{wo/EA} & \makecell[c]{51.49} & \makecell[c]{19.86} & \makecell[c]{18.97} & \makecell[c]{50.18} & \makecell[c]{60.87} & \makecell[c]{46.68} & \makecell[c]{36.07} & \makecell[c]{58.26} & \makecell[c]{74.51} & \makecell[c]{63.60} & \makecell[c]{54.81} & \makecell[c]{70.18} & \makecell[c]{64.42} & \makecell[c]{27.14} & \makecell[c]{25.34} & \makecell[c]{64.02}\\
		\makecell[c]{\multirow{3}{*}{}} & \makecell[c]{CGIR} & \makecell[c]{\textbf{52.03}} & \makecell[c]{\textbf{21.09}} & \makecell[c]{\textbf{19.39}} & \makecell[c]{\textbf{50.55}} & \makecell[c]{\textbf{68.58}} & \makecell[c]{\textbf{49.38}} & \makecell[c]{\textbf{44.69}} & \makecell[c]{\textbf{66.65}} & \makecell[c]{\textbf{76.93}} & \makecell[c]{\textbf{67.16}} & \makecell[c]{\textbf{57.80}} & \makecell[c]{\textbf{72.93}} & \makecell[c]{\textbf{67.54}} & \makecell[c]{\textbf{30.68}} & \makecell[c]{\textbf{29.40}} & \makecell[c]{\textbf{67.29}}\\
		\midrule
		\makecell[c]{\multirow{3}{*}{0.4}} & \makecell[c]{wo/GI} & \makecell[c]{48.04} & \makecell[c]{16.42} & \makecell[c]{14.68} & \makecell[c]{45.73} & \makecell[c]{62.22} & \makecell[c]{41.80} & \makecell[c]{37.16} & \makecell[c]{57.37} & \makecell[c]{74.26} & \makecell[c]{62.21} & \makecell[c]{53.15} & \makecell[c]{67.21} & \makecell[c]{64.66} & \makecell[c]{26.76} & \makecell[c]{25.42} & \makecell[c]{63.15}\\
		\makecell[c]{\multirow{3}{*}{}} & \makecell[c]{wo/EA} & \makecell[c]{48.44} & \makecell[c]{15.98} & \makecell[c]{15.45} & \makecell[c]{\textbf{47.06}} & \makecell[c]{48.75} & \makecell[c]{28.74} & \makecell[c]{21.02} & \makecell[c]{48.07} & \makecell[c]{72.08} & \makecell[c]{61.12} & \makecell[c]{51.11} & \makecell[c]{68.00} & \makecell[c]{61.53} & \makecell[c]{23.10} & \makecell[c]{21.89} & \makecell[c]{60.83}\\
		\makecell[c]{\multirow{3}{*}{}} & \makecell[c]{CGIR} & \makecell[c]{\textbf{49.00}} & \makecell[c]{\textbf{18.39}} & \makecell[c]{\textbf{15.88}} & \makecell[c]{46.56} & \makecell[c]{\textbf{63.82}} & \makecell[c]{\textbf{44.94}} & \makecell[c]{\textbf{38.79}} & \makecell[c]{\textbf{62.67}} & \makecell[c]{\textbf{76.15}} & \makecell[c]{\textbf{65.58}} & \makecell[c]{\textbf{55.77}} & \makecell[c]{\textbf{71.04}} & \makecell[c]{\textbf{65.42}} & \makecell[c]{\textbf{28.35}} & \makecell[c]{\textbf{26.60}} & \makecell[c]{\textbf{64.14}}\\
		\midrule
		\makecell[c]{\multirow{3}{*}{0.6}} & \makecell[c]{wo/GI} & \makecell[c]{43.07} & \makecell[c]{10.96} & \makecell[c]{10.24} & \makecell[c]{40.52} & \makecell[c]{58.50} & \makecell[c]{37.61} & \makecell[c]{32.17} & \makecell[c]{52.94} & \makecell[c]{73.94} & \makecell[c]{61.71} & \makecell[c]{53.29} & \makecell[c]{67.33} & \makecell[c]{63.50} & \makecell[c]{24.67} & \makecell[c]{23.54} & \makecell[c]{62.97}\\
		\makecell[c]{\multirow{3}{*}{}} & \makecell[c]{wo/EA} & \makecell[c]{42.85} & \makecell[c]{10.53} & \makecell[c]{9.79} & \makecell[c]{41.16} & \makecell[c]{39.74} & \makecell[c]{21.65} & \makecell[c]{11.85} & \makecell[c]{36.96} & \makecell[c]{71.80} & \makecell[c]{60.02} & \makecell[c]{50.51} & \makecell[c]{67.83} & \makecell[c]{56.94} & \makecell[c]{17.82} & \makecell[c]{16.68} & \makecell[c]{55.93}\\
		\makecell[c]{\multirow{3}{*}{}} & \makecell[c]{CGIR} & \makecell[c]{\textbf{45.42}} & \makecell[c]{\textbf{15.09}} & \makecell[c]{\textbf{12.48}} & \makecell[c]{\textbf{41.84}} & \makecell[c]{\textbf{59.70}} & \makecell[c]{\textbf{40.10}} & \makecell[c]{\textbf{33.12}} & \makecell[c]{\textbf{57.26}} & \makecell[c]{\textbf{74.78}} & \makecell[c]{\textbf{64.32}} & \makecell[c]{\textbf{53.88}} & \makecell[c]{\textbf{69.93}} & \makecell[c]{\textbf{64.71}} & \makecell[c]{\textbf{26.79}} & \makecell[c]{\textbf{25.53}} & \makecell[c]{\textbf{64.38}}\\
		\bottomrule
	\end{tabular}
	\label{tab:ablation_main}
\end{table*}

\begin{figure}[t]
	\centering
	\begin{subfigure}{0.45\linewidth}
		\includegraphics[width=1\linewidth]{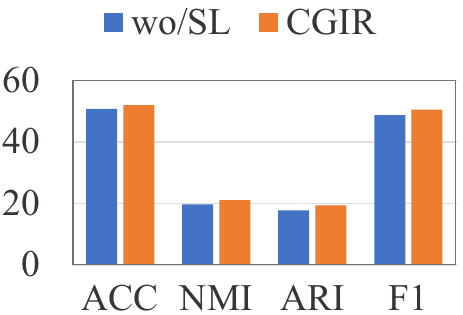}
		\caption{UAT}
	\end{subfigure}
	\hspace{0.02\linewidth}
	\begin{subfigure}{0.45\linewidth}
		\includegraphics[width=1\linewidth]{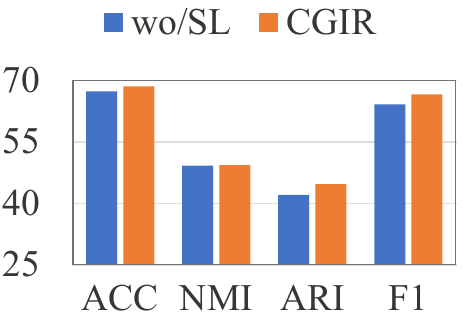}
		\caption{Cora}
	\end{subfigure}
	\caption{Ablation comparison of subcluster-aware loss on two benchmarks with 0.2 missing ratio.}
	\label{fig:ablation_sl}
\end{figure}

Two variants are designed, including wo/GI that suspends the generative imputation and adversarial loss, and wo/EA that removes the edge attention network and contrastive learning. wo/GI performs the KNN-style imputation via the neighbor aggregation mechanism of GCN.
Table \ref{tab:ablation_main} exhibits the ablation results. CGIR still maintains the optimal clustering, which proves the effectiveness of new modules. As the missing ratio increases, the advantage of CGIR over variants becomes salient, further indicating the positive effects of generative imputation and edge attention mechanism on handling attribute-missing graphs. In addition, the variant wo/SL without the subcluster-aware loss is devised that eliminates the subcluster-aware loss, and the ablation comparison is displayed in Fig. \ref{fig:ablation_sl}. wo/SL presents the evident performance degeneration, which reveals that the subcluster-aware loss is beneficial for searching compact subclusters, and impelling accurate imputation and clustering. 

\begin{figure}
	\centering
	\begin{subfigure}{0.3\linewidth}
		\includegraphics[width=1\linewidth]{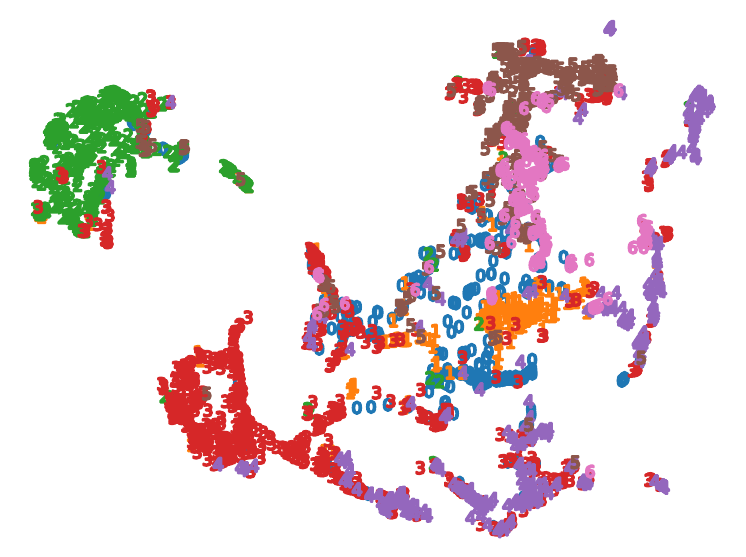}
		\caption{wo/GI}
		\label{fig:ablation_emb1}
	\end{subfigure}
	\hspace{0.02\linewidth}
	\begin{subfigure}{0.3\linewidth}
		\includegraphics[width=1\linewidth]{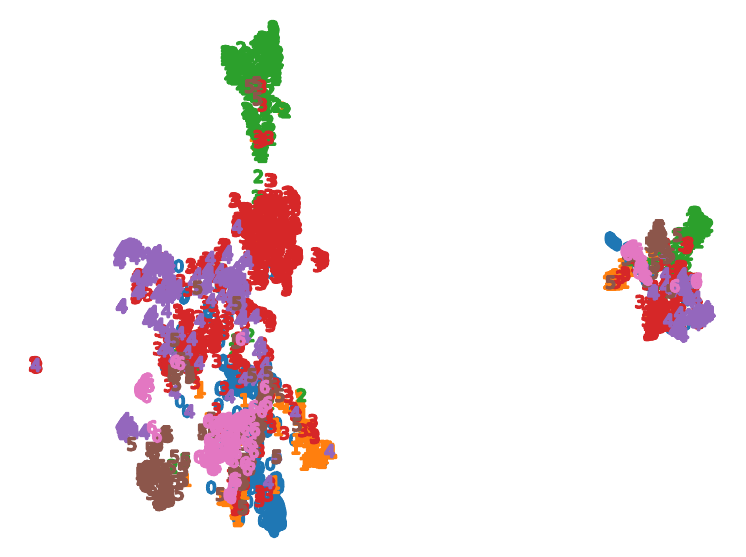}
		\caption{wo/EA}
		\label{fig:ablation_emb2}
	\end{subfigure}
	\hspace{0.02\linewidth}
	\begin{subfigure}{0.3\linewidth}
		\includegraphics[width=1\linewidth]{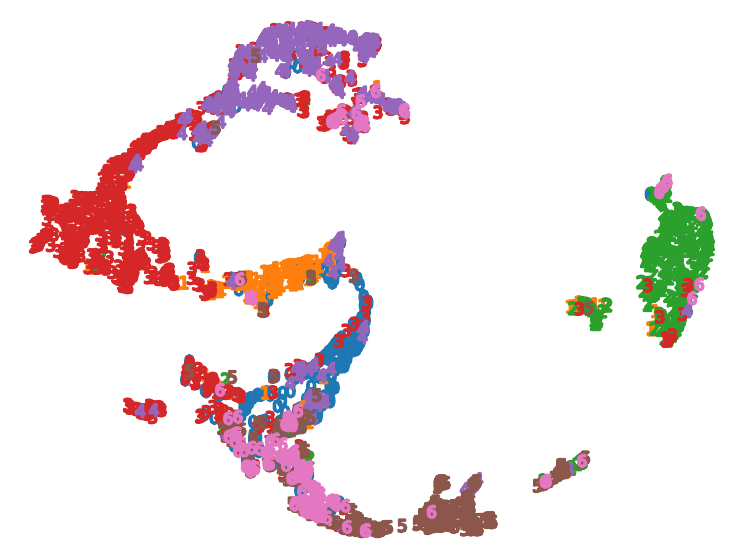}
		\caption{CGIR}
		\label{fig:ablation_emb3}
	\end{subfigure}
	\caption{Visualization of learned embedding $\mathbf{F}$ on Cora with 0.2 missing ratio.}
	\label{fig:ablation_emb}
\end{figure}

To illustrate the effects of new modules intuitively, the output embedding is visualized by UMAP \cite{umap} in Fig. \ref{fig:ablation_emb}. Compared with wo/GI, the embedding learned by CGIR presents the more distinguishable clusters, which manifests that CGIR can estimate the subcluster distributions accurately to implement clustering-friendly imputation. The imputation nodes possess class-specific semantic information to retain compact with intra-class nodes. Additionally, the superiority of CGIR against wo/EA indicates that EAN can detect the edge-wise attributes to execute more reliable structure recovery. The visualization comparison again shows that the presented new modules can suppress the adverse effects of attribute absence.

\begin{figure}
	\centering
	\begin{subfigure}{0.46\linewidth}
		\includegraphics[width=1\linewidth]{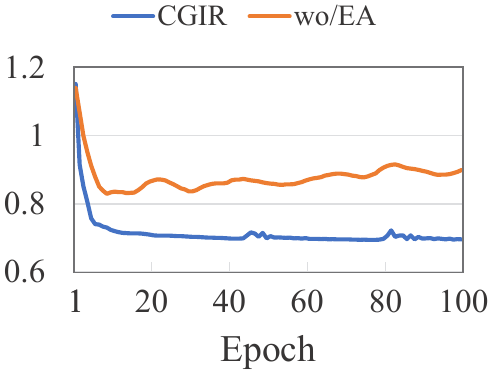}
		\caption{UAT}
	\end{subfigure}
	\hspace{0.02\linewidth}
	\begin{subfigure}{0.46\linewidth}
		\includegraphics[width=1\linewidth]{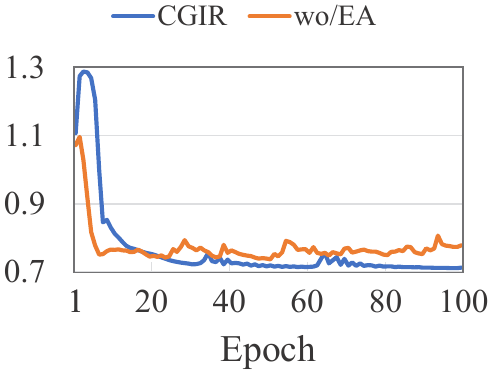}
		\caption{Cora}
	\end{subfigure}
	\caption{Convergence trend of graph reconstruction loss ${\mathcal{L}}_{gra}$ on two benchmarks with 0.2 missing ratio.}
	\label{fig:convergence}
\end{figure}

The convergence trend of graph reconstruction loss ${\mathcal{L}}_{gra}$ is also plotted to demonstrate the advantage of EAN. As shown in Fig. \ref{fig:convergence}, CGIR achieves a lower reconstruction residue compared with wo/EA. The experimental result further testifies that EAN can learn edge-wise attributes to predict the initial graph structure more precisely, so as to provide reliable refinement for the proposed clustering-oriented generative attribute graph imputation.

\subsection{Parameter Sensitivity}

\begin{figure}
	\centering
	\begin{subfigure}{0.45\linewidth}
		\includegraphics[width=1\linewidth]{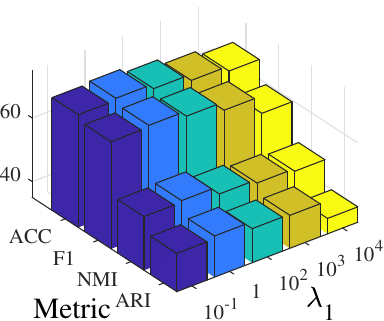}
		\caption{$\lambda_1$}
		\label{fig:xxx}
	\end{subfigure}
	\hspace{0.02\linewidth}
	\begin{subfigure}{0.45\linewidth}
		\includegraphics[width=1\linewidth]{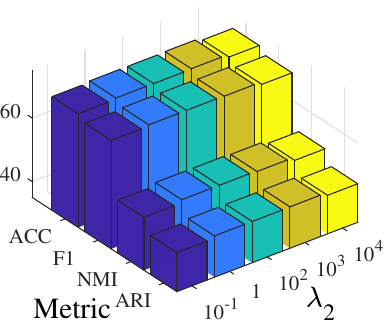}
		\caption{$\lambda_2$}
	\end{subfigure}
	\begin{subfigure}{0.45\linewidth}
		\includegraphics[width=1\linewidth]{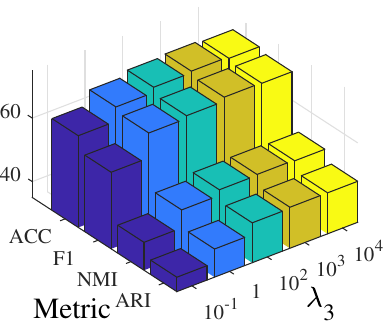}
		\caption{$\lambda_3$}
	\end{subfigure}
	\hspace{0.02\linewidth}
	\begin{subfigure}{0.45\linewidth}
		\includegraphics[width=1\linewidth]{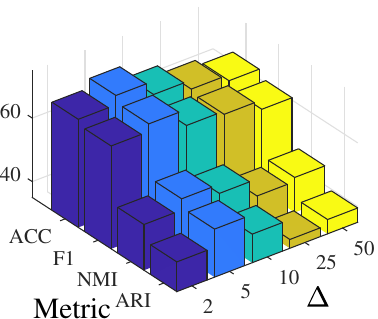}
		\caption{$m = c \times \Delta$}
	\end{subfigure}
	\caption{Parameter sensitivity of CGIR on Cora with 0.2 missing ratio. }
	\label{fig:sensitivity}
\end{figure}

Finally, the sensitivity of CGIR to hyper-parameter selection is investigated. To precisely reflect the impact of each hyper-parameter, we only adjusts one hyper-parameter in each analysis, and fix the remaining. Fig. \ref{fig:sensitivity}(a-c) shows the performance fluctuation under different trade-off parameters $\lambda_1$, $\lambda_2$, and $\lambda_3$. It is observed that the influence of $\lambda_1$ and $\lambda_3$ is more prominent than that of $\lambda_2$, because $\lambda_1$ and $\lambda_3$ are directly related to clustering-oriented generative imputation and edge-attentional refinement, respectively. The performance is relatively steady within an appropriate range. In addition, Fig. \ref{fig:sensitivity}(d) displays the metric scores under distinct subcluster number $m$. When $m$ is too large, it is difficult to capture sufficient class-specific semantic information to support an accurate imputation for clustering. The setting of a moderate subcluster number (e.g., $m = c \times 5$) tends to bring a satisfactory result.

\section{Conclusion}

In this paper, we propose a new GCN-based framework called CGIR for attribute-missing graph clustering. 
For node completion, CGIR searches compact subcluster distributions, and estimates the missing attributes through a generative adversarial scheme. In this way, the class-relevant characteristics are preserved to guide clustering-oriented imputation. 
Besides, the edge-wise attributes are determined according to the link relationship of nodes, which impels more accurate graph reconstruction to support reliable imputation refinement.
Extensive experiments on several benchmarks demonstrate the superior clustering performance of CGIR on handling attribute-missing graphs. 

\section*{Acknowledgments}

This work was supported by the National Key Research and Development Program of China (Grant No: 2022ZD0160803).

\bibliographystyle{ACM-Reference-Format}
\bibliography{sample-base}


\end{document}